\documentclass{article}

\PassOptionsToPackage{numbers, compress}{natbib}

\usepackage[final]{nips_2018}

\usepackage[utf8]{inputenc} 
\usepackage[T1]{fontenc}    
\usepackage{hyperref}       
\usepackage{url}            
\usepackage{booktabs}       
\usepackage{amsfonts}       
\usepackage{nicefrac}       
\usepackage{microtype}      
\usepackage{subfigure} 
\usepackage{graphicx} 

\usepackage{bm}
\usepackage{amsmath}
\usepackage{amsthm}
\usepackage{pdfpages}

\usepackage{xcolor}
\newcommand\anote[1]{\textcolor{red}{#1}} 

\renewcommand{\b}{\mathbf}

\newcommand{\bb}[1]{\mathbf {\bar {{#1}}}}
\newcommand{\Ln}{\left \Vert}
\newcommand{\Rn}{\right \Vert}

\title{Towards Global Remote Discharge Estimation:\\Using the Few to Estimate The Many}

\renewcommand*{\thefootnote}{\fnsymbol{footnote}}

\author{
  \textbf{Yotam Gigi}^{\ref{gr_huji}}\ \ \textbf{Gal Elidan}^{\ref{gr_huji}}\ \  \textbf{Avinatan Hassidim}^{\ref{gr_bi}}\ \ \textbf{Yossi Matias}^{\ref{gr}}\ \  \\
  \textbf{Zach Moshe}^{\ref{gr}}\ \   \textbf{Sella Nevo}^{\ref{gr}}\ \ \textbf{Guy Shalev}^{\ref{gr}}\ \ \textbf{Ami Wiesel}^{\ref{gr_huji}}
}

\begin{document}

\maketitle

\begin{abstract}
Learning hydrologic models for accurate riverine flood prediction at scale is a challenge of great importance. One of the key difficulties is the need to rely on in-situ river discharge measurements,  which can be quite scarce and unreliable, particularly in regions where floods cause the most damage every year. Accordingly, in this work we tackle the problem of river discharge estimation at different river locations.
A core characteristic of the data at hand (e.g. satellite measurements) is that we have few measurements for many locations, all sharing the same physics that underlie the water discharge. We capture this scenario
in a simple but powerful common mechanism regression (CMR) model with a local component as well as a shared one which captures the global discharge mechanism. The resulting learning objective is non-convex, but we show that we can find its global optimum by leveraging the power of joining local measurements across sites. In particular, using a spectral initialization with provable near-optimal accuracy, we can find the optimum using standard descent methods. We demonstrate the efficacy of our approach for the problem of discharge estimation using simulations.  
    \footnotetext[2]{\label{gr_huji} Google Research and The Hebrew University of Jerusalem, Israel}
    \footnotetext[3]{\label{gr_bi} Google Research and Bar-Ilan University}
    \footnotetext[4]{\label{gr} Google Research} 
\end{abstract}

\renewcommand*{\thefootnote}{\arabic{footnote}}

\section{Introduction}
Floods are the most common and deadly natural disaster in the world. Every year, floods cause between thousands to tens of thousands of fatalities \cite{cred, jonkman2003loss, unisdr, jonkman2005global, doocy2013human}, affect hundreds of millions of people \cite{doocy2013human, jonkman2005global, unisdr}, and cause tens of billions of dollars in economic damages \cite{cred, unisdr}. Sadly, these numbers have only been increasing in recent decades \cite{loster1999flood}. Indeed, the UN charter notes floods to be one of the key motivators for the formulation of the sustainable development goals (SDGs), and directly challenges us: "They knew that earthquakes and floods were inevitable, but that the high death tolls were not" \cite{undp}.

Early warning systems, even with limited lead time and imperfect accuracy, 
have been shown to reduce both fatalities and economic damages by more than a third, and in some cases almost by half \cite{who, pilon1998guidelines, worldbank}. Unfortunately, the majority of human costs that are due to flooding are concentrated in developing countries \cite{doocy2013human}, which often lack effective and actionable early warning systems due to limited data collection, funding, or professional expertise \cite{stromberg2007natural}. The result is that, across multiple countries, thousands die on average every year, and relief and mitigation efforts have very limited information to rely on.

In this work, as part of our broader efforts in flood forecasting \cite{nevo2018ml}, we focus on riverine floods which are responsible for much of the effect on human life.
Existing hydrologic methodology for building flood prediction models relies heavily on in-situ infrastructure such as costly extensive gauging systems \cite{worldbank}, and on local adaption of the models that requires highly trained professionals \cite{anderson2002calibration}. Providing value where it matters most thus requires overcoming several challenges. First, we would like to reduce reliance on in-situ measurements such as extensive gauging sites constructed along the modeled river. Relevant data is constantly being produced at immense scale across the globe, but the vast majority of this data is not measured using in-situ measurements but rather comes in the form of, e.g., satellite imagery. Clearly, leveraging even small parts of it has the potential for substantially improving flood prediction models. Second, to cover large areas in developing regions, we must automate and scale up the model building methodology and reduce its reliance on the human factor. Third, it is the sad paradox of life that populations in low-means areas cannot afford to respond to a low precision system, and thus to make positive impact in such areas, we require improved predictive power.

The field of machine learning (ML) has transformed many aspects of our lives, and is naturally geared to cope with the above challenges. Improving prediction, leveraging on multiple signals that are difficult for a human expert to get a grasp on, and automating human processes, are all characteristics of effective ML systems. The first critical step toward building such systems is to provide global-scale estimates of the water discharge (volume per second) through the cross sections of a river, which can then be used to train early warning predictive models. As noted, such in-situ measurements are unavailable more often than not, and thus our first goal is to perform remote discharge estimation, or estimating the discharge based on remote measurements (usually satellite data) \cite{smith1997satellite}.


Our concrete goal is thus to create a prediction model that, using few measurements from a set of river locations, will be able to generalize to \emph{all} locations. Intuitively, this should be possible because the multiple prediction problems (one for each location) are related: the underlying physical mechanism that relates satellite measurements to water level is identical, and each local measurement, where it exists, gives us a "clue" as to the nature of this shared mechanism.
This general setting of leveraging information about some tasks to assist in the learning of models for other tasks has a long history in machine learning: inductive transfer, transfer learning and multitask learning are all closely related variants of the framework (see, e.g. \cite{Thrun:1996, InductiveTransfer:1997, Caruana:1997, Baxter:2000} for some of the early influential works and \cite{Pan:2010} for a more recent survey). 


We consider a simple but powerful regression model where the coefficients are composed of two components: one local that allow us to adapt to the characteristics of the local site, and one shared that allows us to capture the global water discharge mechanism. It is this shared component that can benefit from transfer learning. A similar formal setting is explored in the highly cited work of \cite{ando2005framework} where the task is called \emph{structural learning}%
\footnote{The focus of the work on transferring from unlabeled to labeled tasks is different from ours but the formal underpinning is identical.}, pointing to the common shared structure learned. As they show, using the empirical risk minimization (ERM) principle, it is provably beneficial to learn from multiple tasks, from a statistical sample complexity perspective. 
A recent work \cite{yuan2017spectral} also shows \emph{empirically} that this shared regression approach can be useful for multispectral imagery classification.
The computational and optimization questions of \emph{"Can we efficiently learn such a model?"} are left unanswered. In this work, we show that the answer to this question, at least from an optimization viewpoint, is in the affirmative.

The target objective of this common mechanism regression (CMR) is non-convex and may have spurious local minima. Our main contribution is that, given enough independent tasks, we can efficiently find its global optimum. For this purpose, we extend the ideas in \cite{netrapalli2013phase,candes2015phase} to CMR with multiple regressions. We begin with a spectral initialization with provable near-optimal accuracy, and then refine it using standard descent methods. 


In the context of remote discharge estimation, our learning goal is to capture the common discharge mechanism that relates satellite measurements from multiple spectral bands to water levels. Naturally, we do not have access to the true mechanism (or we would not need to learn it). However, we can simulate such mechanisms and assess the merit of our approach when the ground truth is known. Using such simulations, we demonstrate the effectiveness of using our approach for transfer learning: sharing measurements from individual sites allow us to jointly improve the average predictive performance across all of them.
\section{Common Mechanism Regression (CMR)}
Our model consists of $I$ independent regressions that share a common mechanism. For simplicity, we assume that each regression has exactly $T$ pairs of labels and features 
\begin{eqnarray}\label{data}
    \left\{y_{it},\b{X}_{it}\right\}_{t=1}^T \quad i=1,\cdots,I
\end{eqnarray}
where $y_{it}$ are scalar labels, and $\b{X}_{it} \in \mathbb R^{B \times P}$ are matrix observations.%
\footnote{We use $B$ for bands and $P$ for pixels in the context of discharge estimation but the setting is general.} 
Our common mechanism regression (CMR) involves a two phase approach: a common mechanism parameterized by $
\b w$ followed by decoupled local linear regressions denoted by $\b v_i$:
\begin{eqnarray}
    \label{model}  y_{it} = \b w^T\b X_{it} \b{v}_i.
\end{eqnarray}
Note that the overall structure is linear in the features, but has a bilinear parameterization.
Our main goal is to recover the common parameter $\b{w}$ and, if possible, we would also like to identify the local $ \b{v}_i $'s. In particular, we are interested in the scenario when $I$ is large but $T$ is small, so that we have many regression problems but few observations for each one. Each regression, if estimated independently, requires at least $T>B+P$ samples. By introducing a common mechanism where $\b w$ is shared across the different sites, we allow $P<T<B+P$, and also address the case of $T<P$ where exact recovery of $\b{v}_i$ is impossible.

The CMR model is natural for river discharge estimation using remote sensing. Specifically, in multispectral imaging, the data matrices $\b{X}_{it}$ are defined by spectral and spatial dimensions. 
A reasonable approach to discharge estimation is thus to use the spectral information to identify water pixels and then apply spatial regression. The classical technique for water identification is via a {\emph{common}} non-linear spectral feature known as Normalized Difference Water Index (NDWI) \cite{mcfeeters1996use}\footnote{More advanced indices are reviewed in \cite{isikdogan2017surface}.}. This index is the motivation to CMR which automatically learns a data-driven feature defined by the weights of $\b w$.
In what follows, we will show that linear CMR outperforms the non-linear NDWI.

We propose to recover the parameters as the solution to the following regularized bilinear least squares optimization:
\begin{equation} \label{wv_opt_problem}
(CMR):\quad \min_{\b{w} \in \mathbb R^B , \\ \b {v}_i \in \mathbb R^P} 
    \sum_{i,t} \left( y_{it} - \b{w}^T \b{X}_{it} \b{v}_i \right)^2 + \lambda\sum_i\|\b{v}_i\|^2\quad {\rm{s.t.}} \quad \|\b{w}\|=1
\end{equation}
Due to its bi-linear structure, CMR involves a non-convex minimization. Naive descent techniques may therefore converge to spurious local minima. Interestingly, CMR is similar to phase retrieval problems where it was recently shown that these bad critical points can be avoided via clever initialization schemes \cite{netrapalli2013phase,candes2015phase}. Adaptation of these ideas to CMR leads to the following {\em{common}} spectral initialization:
\[
\b Z_i = \frac 1T \sum_t y_{it} \b X_{it}, \qquad
\b Q =\frac{1}{I} \sum_i^I \b Z_i \b Z_i^T,
\qquad
\b w^0 = {\rm {eigv}}_1 \left( \b Q \right)
\]
where $\rm {eigv}_1$ is the eigenvector corresponding to the largest eigenvalue. 
From here, we continue with standard descent methods, e.g., gradient descent or alternating least squares, till convergence.
 Together, the computational complexity of this approach is linear in $I$.  


Under standard assumptions, 
the proposed spectral initialization can recover the true $\bb{w}$ with high accuracy. Like \cite{hardt2016identity}, we consider the  realizable case, with normal features and assume an exact CMR model with no noise. We also assume random local regressors, i.e., we model $\bb {v}_i$ as i.i.d. realizations of an arbitrary probability distribution. This last assumption is special for our work and is required in order to model multiple regression problems with common characteristics. 



{\bf{Theorem}} {\em{
Under the above assumptions, there exists a constant $C>0$ such that if $\sqrt{IT}\geq CP^2B^2/\epsilon^2$, then $\text{dist}(\b w^0, \bb w)\leq \sqrt{\epsilon}$ 
with probability of at least $1 - \frac 1{\sqrt{IT}}$.
}}

The theorem quantifies the improved performance when increasing $T$ or $I$ via their product. To prove the theorem, we show that
\begin{eqnarray}
    \mathbb E[\b Q]=\alpha\bb w \bb w^T+\beta \b I 
\end{eqnarray}
where $\alpha$ and $\beta$ are positive constants that depend on the distribution of $\b v_i$. Thus, its principal eigenvector is the true parameter. Using the fact that the variance of $\b Z_i$ decays with $T$, we show that $\b Q$ concentrates around its mean as $I$ and $T$ increase.

\section{Numerical experiments}


We start by assessing the merit of our CMR approach for discharge estimation using synthetic simulations. Recall that our goal is to leverage measurements from many locations to improve prediction. Thus, we consider the performance of CMR for a range of values of $I$
(the number of sites) and $T$ (the number of samples per site). For each set of $I, T$ values, we repeat the following 50 times: chose a random $\b w$ and $\b v_i$, run the CMR algorithm, and declare success if the squared correlation between the true $\b w$ and its estimate exceeds $0.90$. 
We do this with and without the spectral initialization. The results for $B=20$ and $P=10$ are summarized in the figure below. 

\begin{figure}[h]
\includegraphics[trim=0 0 100 0, clip,height=0.31\textwidth]{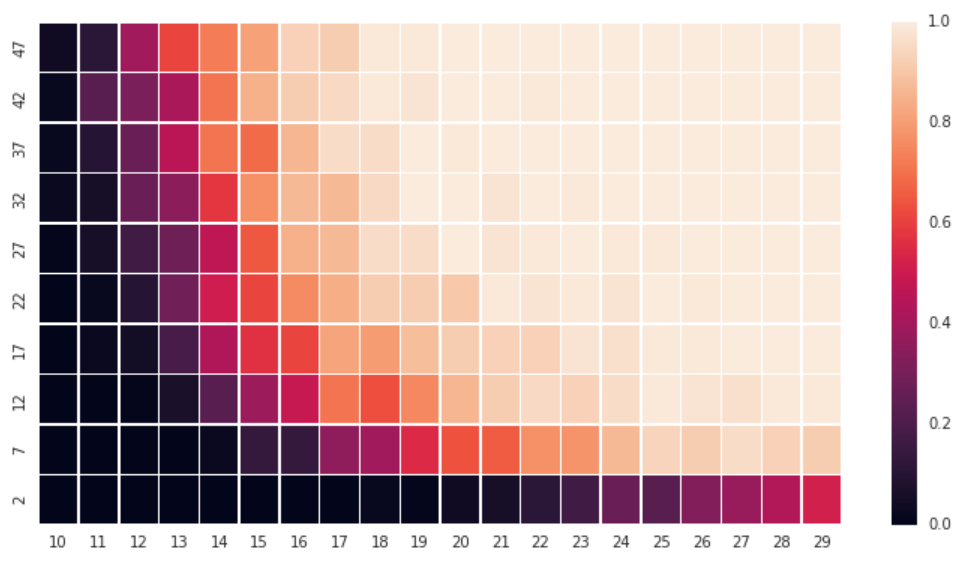}
\includegraphics[height=0.31\textwidth]{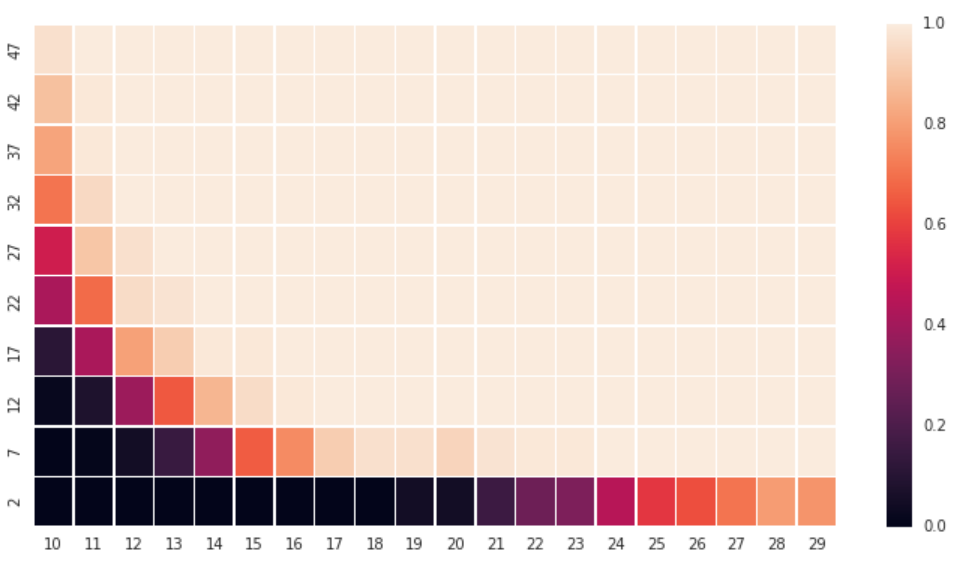}

\caption{Recovery of the true shared mechanism $\b w$ using the CMR model as a function of the number of sites $I$ (y-axis) and the number of samples per site $T$ (x-axis) without (left) and with (right) spectral initialization. The color of each square corresponds to the fraction of successful recoveries.}
\label{success_graphs}
\vspace{0.1in}
\end{figure}

As expected, the results demonstrate that CMR recovers $\b w$ with few samples for many sites, i.e., when  $I>B>T>P$. Interestingly, we also succeed in recovering $\b w$ when $T<P$, a setting where it is impossible to recover $\b v_i$. The left and right panels illustrate the importance of the  initialization, which substantially widens the ranges of settings for which CMR succeeds with high probability.

We now evaluate the merit of our CMR approach for the predictive task of discharge estimation in a real-world setting. We use images from LANDSAT8 mission \cite{roy2014landsat} which include 11 spectral bands each, and ground truth labels from the United States Geological Survey (USGS) website.
The results were generated using $23$ river gauge sites with $100$ temporal samples each. For every cross validation fold, the temporal samples were split into train and test, and the CMR results were compared with the NDWI per-site regression. The average mean squared errors, normalized per-site, of randomly shuffled $4$-fold cross validation repeated 4 times are given in the following table:

\begin{center}
    \begin{tabular}{|c|c|c|c|c|}
\hline
 & Train & Test \\
 \hline
   NDWI & 0.54 & 0.70 \\
    CMR & 0.47 & 0.65 \\
    \hline
\end{tabular}
\end{center}

As can be seen, there is a clear advantage to learning the shared component of the CMR model. Appealingly, the advantage is also substantial on held out test data, despite the expressiveness of the CMR model which also allows for local components.

\section{Summary and Future Directions}
In this work, we proved that, despite the non-convex nature of the learning objective, the common mechanism regression (CMR) model can be globally optimized using a spectral initialization combined with standard descent. We also demonstrated the efficacy of the approach for the challenge of discharge estimation where we have few measurements for many river sites.

On the modeling front, it would be useful to generalize CMR so as to allow for robust and task-normalized loss functions. Another interesting direction is to inject non-linearity into CMR to make it even more competitive with the non-linear physically motivated NDWI approach. On the practical discharge estimation front, we plan to aggregate multiple data sources (e.g. additional types of satellites, weather data) within the CMR framework.





\bibliography{bib_aistats}
\bibliographystyle{plain}

\end{document}